\documentclass{article}
\usepackage{amssymb}
\usepackage{xcolor}
\usepackage[preprint]{corl_2025}

\usepackage{amsmath,amsfonts}
\usepackage{bm}
\usepackage{bbm}
\usepackage{graphicx}
\usepackage{booktabs}
\usepackage{multirow}
\usepackage{multicol}
\usepackage{array}
\usepackage{calc}
\usepackage{wrapfig}
\usepackage{subfigure}
\usepackage{algorithm}
\usepackage{algpseudocode}
\usepackage{appendix}

\newcolumntype{L}[1]{>{\raggedright\arraybackslash}p{#1}}

\algrenewcommand{\algorithmicrequire}{\textbf{Input:}}
\algrenewcommand{\algorithmicensure}{\textbf{Output:}}

\title{Pandora: Articulated 3D Scene Graphs from Egocentric Vision}

\author{
  Alan Yu\textsuperscript{1} \quad
  Yun Chang\textsuperscript{1} \quad
  Christopher Xie\textsuperscript{2} \quad
  Luca Carlone\textsuperscript{1} \\
  \textsuperscript{1}Laboratory for Information and Decision Systems,
  Massachusetts Institute of Technology \\
  \textsuperscript{2}Meta Reality Labs \\
  \texttt{\{alanyu, yunchang, lcarlone\}@mit.edu},
  \texttt{chrisdxie@meta.com}
}

\begin{document}
\maketitle
\vspace{-0.2in}

\begin{abstract}
Robotic mapping systems typically approach building metric-semantic scene representations from the robot's own sensors and cameras. However, these ``first person'' maps inherit the robot's own limitations due to its embodiment or skillset, which may leave many aspects of the environment unexplored. For example, the robot might not be able to open drawers or access wall cabinets. In this sense, the map representation  is not as \textit{complete}, and requires a more capable robot to fill in the gaps. We narrow these blind spots in current methods by leveraging egocentric data captured as a human naturally explores a scene wearing Project Aria glasses~\citep{engel2023projectarianewtool}, giving a way to directly transfer knowledge about articulation from the human to any deployable robot. We demonstrate that, by using simple heuristics, we can leverage egocentric data to recover models of articulate object parts, with quality comparable to those of state-of-the-art methods based on other input modalities. We also show how to integrate these models into 3D scene graph representations, leading to a better understanding of object dynamics and object-container relationships. We finally demonstrate that these articulated 3D scene graphs enhance a robot's ability to perform mobile manipulation tasks, showcasing an application where a Boston Dynamics Spot is tasked with retrieving concealed target items, given only the 3D scene graph as input.
\end{abstract}


\section{Introduction}
\label{sec:intro}

\begin{figure}[t]
  \centering
  \includegraphics[width=0.85\linewidth]{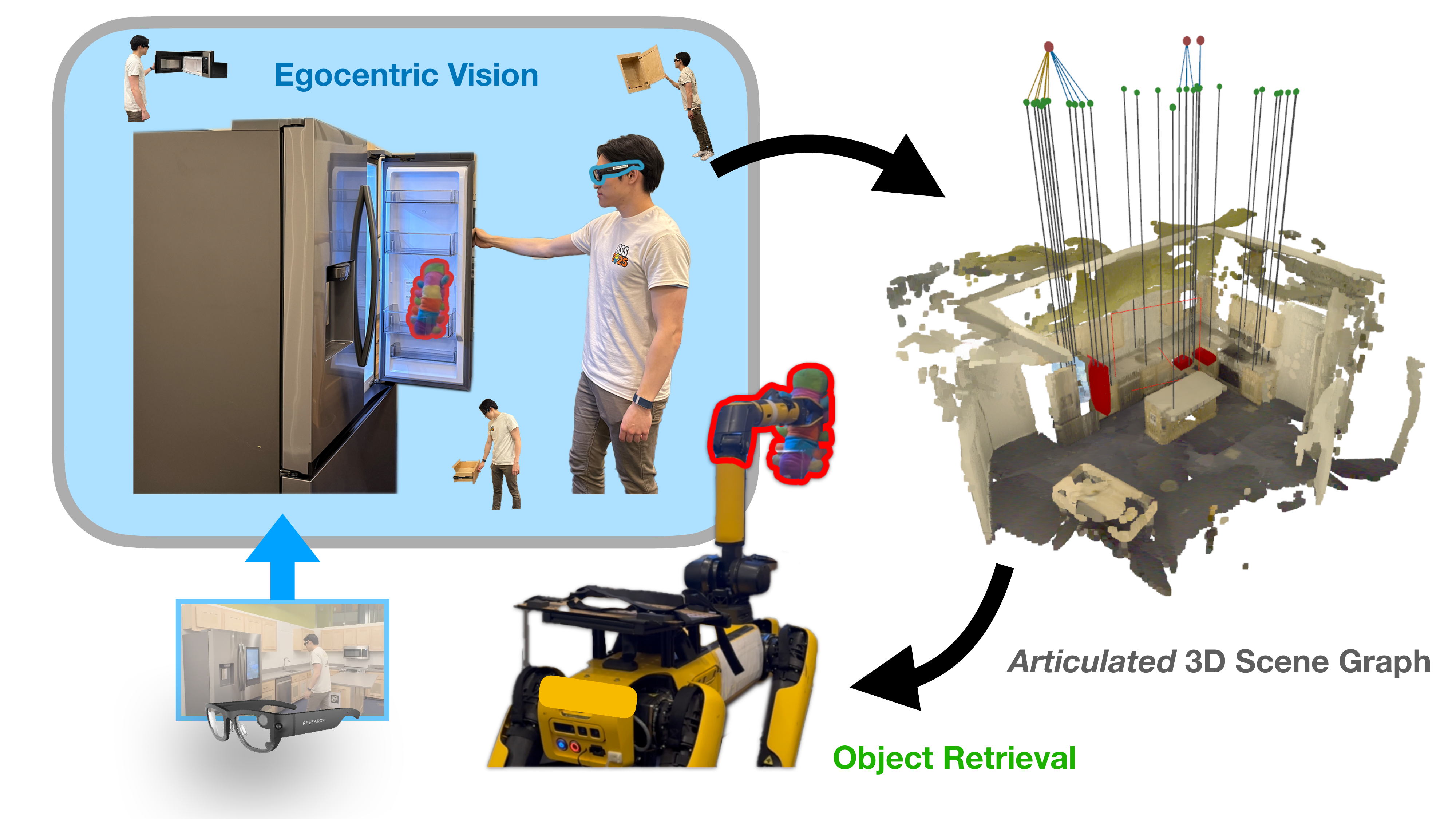}
  \caption{Pandora constructs an \textit{articulated} 3D scene graph from egocentric data, where we model articulate object parts and their relationships to objects they contain. The estimated articulation models from human interactions are used to build a 3D scene graph, which can be then used for downstream object-retrieval tasks on a mobile robot.}
  \label{fig:teaser}
\end{figure}
Understanding how objects move ---and how they are interacted with--- is essential for enabling robots to operate effectively in unstructured, human-centric environments. Everyday tasks such as finding a lost item, restocking a fridge, or tidying a living room rely not just on recognizing objects, but on modeling their articulations: how doors swing, drawers slide, and containers conceal their contents. These interactions are inherently dynamic, requiring representations that capture both what objects are and how they can change state. To reason over such complex, dynamic environments, 3D scene graphs offer a compelling structure: they provide a hierarchical representation of a scene's objects, their spatial layout, and inter-object relationships.
By enriching these graphs with articulation models,
we can support better reasoning capabilities for robotics ---enabling more accurate representation of interobject relationships, affordances, and feasible manipulation strategies.

In this work, we argue that egocentric data ---captured from humans naturally interacting with their environment--- offers a rich and underutilized signal for constructing such representations. Unlike robot-centric approaches, which suffer from limited viewpoint, embodiment constraints, and restricted articulation capabilities, egocentric recordings provide broader spatial coverage and encode implicit interaction labels (e.g., hand poses and trajectories) that reveal how objects move and function.

We introduce Pandora, a method for building an \textit{articulated} 3D scene graph directly from egocentric data captured with Project Aria glasses~\citep{engel2023projectarianewtool} (Figure~\ref{fig:teaser}), that uses estimated articulation models to understand concealed objects. We utilize this representation by deploying a Boston Dynamics Spot in an object-retrieval setting, relying only on the representation to inform its navigation and actions.
Our contributions include (1) a simple method for estimating articulation models from egocentric vision; (2) integrating them into \textit{articulated} scene graphs that encode object-container relationships, dynamics, and articulation; and (3) a direct application of our system in a real-world object-retrieval setting.

\section{Related Work}

\paragraph{3D Scene Graphs}
Early work in 3D scene graphs represented scenes as objects, rooms, and buildings with relational edges \citep{armeni20193d}. Systems like Hydra~\citep{hughes2022hydra, chang2023hydramulticollaborativeonlineconstruction} and Khronos~\citep{Schmid-RSS24-Khronos} have been successful at constructing 3D scene graphs in real time, with Khronos adding temporal understanding by tracking short-term object centroid movement and detecting long-term changes. Other works enrich scene graphs with affordances~\citep{zhang2025open}, open-vocabulary semantics~\citep{conceptgraphs, Werby2024hovsg}, and robot exploration actions~\citep{jiang2024roboexp}, but do not explicitly address articulation. Object relations have been previously studied in~\citep{Kim_2020, wald2020learning3dsemanticscene, wu2021scenegraphfusionincremental3dscene}, but their kinematics are not accounted for. We instead provide the scene graph with additional structure by modeling articulate objects.

\paragraph{Articulation Model Estimation}
Probabilistic approaches model an articulate object as a kinematic graph, but rely on privileged information about the part-level poses throughout interaction~\citep{sturm2008icra, sturm2008rss, sturm2009ijcai, sturm2011jair}. More recently, learning-based approaches use standard onboard sensor modalities to infer these models. Methods like ScrewNet~\citep{jain2021screwnetcategoryindependentarticulationmodel} and DUST-net~\citep{jain2021distributionaldepthbasedestimationobject} utilize screw theory to estimate articulation parameters from depth. FlowBot3D~\citep{eisner2024flowbot3dlearning3darticulation} predicts flow vectors representing articulation and utilize the predictions with a robot manipulator. Ditto~\citep{jiang2022dittobuildingdigitaltwins} and DiTH~\citep{Hsu2023DittoITH} additionally reconstruct part-level geometry, which is used by Kinscene~\citep{hsu2024kinscenemodelbasedmobilemanipulation} to articulate scenes to arbitrary configurations with a mobile robot. As the exploration is done entirely by the robot, the resulting scene representation inherits many of the limitations previously discussed. More recently, Articulate Anything~\citep{le2024articulate} leverages vision-language priors for articulation model estimation.

\paragraph{Egocentric Vision for Scene Understanding}
Egocentric data offers a natural vantage point on human--scene interactions \citep{adt} and everyday behavior \citep{Damen2018EPICKITCHENS, grauman2022ego4dworld3000hours}, revealing subtle cues like hand trajectories, eye gaze, and object affordances that remain difficult to discern from third-person or robot observations. EgoGaussian~\citep{zhang2024egogaussian} leverages egocentric video data to enable dynamic scene reconstruction through a combination of 2D hand segmentation and 3D Gaussian Splatting. Similarly, LostFound~\citep{behrens2024lostfoundupdating} uses hand tracking information from the Aria device~\citep{engel2023projectarianewtool} to update an underlying scene graph representation from human-object interaction events. Prior work also addresses grounding 3D affordances~\citep{liu2024grounding3dsceneaffordance}, but there is still a gap in the direction of using egocentric data to incorporate articulation into scene reconstructions.

\section{Pandora}
\label{sec:scene-graph}

We consider a setting where a human, equipped with an egocentric device such as the Project Aria glasses~\citep{engel2023projectarianewtool}, explores a scene and collects a sequence of observations $\{\mathbf{S}_t\}_{t=1}^N$. Each observation $\mathbf{S}_t$ at the time $t$ consists of hand pose $\mathbf{H}_t \in \mathbb{R}^3$, RGB image $\mathbf{I}_t \in \mathbb{R}^{H\times W \times 3}$, depth image $\mathbf{D}_t \in \mathbb{R}^{H\times W}$, and camera pose $\mathbf{T}_t \in SE(3)$.
During the scan, the human articulates a set of object parts, where the $i$-th object part is defined by its geometry $\mathbf{P}_i$ and joint parameters $\mathbf{A}_i$.
We represent each $\mathbf{P}_i$ as a pointcloud. Each $\mathbf{A}_i$ consists of the articulation type $\kappa \in$ \{Prismatic, Revolute\}, along with the axis $\mathbf{u}$ in the prismatic case, or the axis $\mathbf{u}$ and pivot $\mathbf{x}$ in the revolute case. Due to the lack of a depth sensor on the Aria device, we implicitly obtain depth as described in Appendix~\ref{appendix:inferring-depth}. We assume that when an object part is first articulated, the operator captures the part in its entirety to allow for complete reconstruction. From these observations, we seek to reconstruct an \textit{Articulated 3D Scene Graph}, as defined below.

\subsection{Articulated 3D Scene Graphs}
We define an \textit{Articulated 3D Scene Graph} to consist of a hierarchical graph $\mathcal{G}$ that describes the geometry and relationship of the objects in the scene, and an event mapping $\mathcal{E}$ that describes the articulation states across time.

\begin{wrapfigure}[18]{R}{0.6\textwidth}
  \vspace{-1.5\baselineskip}
  \centering
  \includegraphics[width=0.95\linewidth]{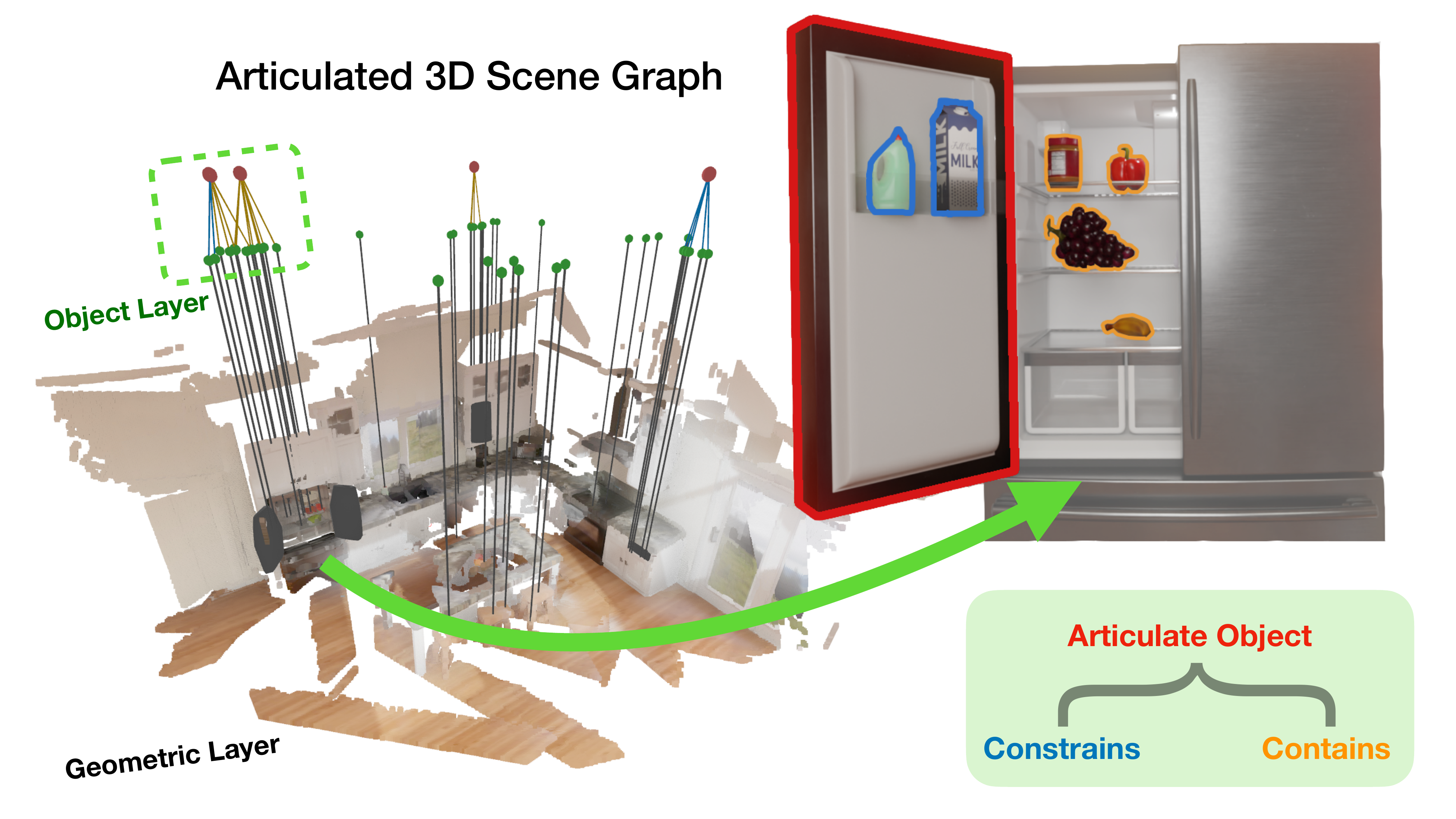}
  \caption{Pandora models the \textit{Geometric Layer}, represented by a 3D voxel grid, and the \textit{Object Layer}. Objects can be either (1) articulate parts or (2) ordinary objects. Articulate parts (e.g.\ the fridge door) can \textit{constrain} the movement of ordinary objects (e.g.\ the soap bottle and milk carton on the door itself), or only \textit{contain} them (e.g.\ the grapes that lie behind the door).}
  \label{fig:scene-graph-structure}
\end{wrapfigure}

\paragraph{Scene Graph} As illustrated by Figure~\ref{fig:scene-graph-structure}, $\mathcal{G}$ consists of two layers: The \textit{Geometric Layer}, which is a 3D voxel grid representing the geometry of the scene, and the \textit{Object Layer}, which is a graph $(\mathbf{V}, \mathbf{E})$ of objects, where each object $v_i \in \mathbf{V}$ can be one of two types: (1) an articulated part, annotated with joint parameters $\mathbf{A}_i$; or (2) a static object, annotated with a CLIP~\citep{radford2021learningtransferablevisualmodels} feature. An edge $(v_i, v_j) \in \mathbf{E}$ in the objects layer connects an articulated part to a child static object and encodes their relationship as \textit{constrains}, indicating that $v_i$ constrains the motion of $v_j$ (e.g., cutlery inside a drawer follow the drawer's translation), or \textit{contains}, which indicates that $v_j$ is contained, but not constrained, by $v_i$ (e.g., plates sitting behind a cabinet door).
Interlayer edges connect each object node with voxels in the \textit{Geometric Layer} (for articulate objects, this is exactly $\mathbf{P}_i$). We denote with ${\mathcal{A}}$ and ${\mathcal{P}}$ the sets of joint parameters and point clouds for all articulated parts observed in the scene, respectively.

In this work, we primarily focus on modeling these two layers, but note that it is possible to extend to higher-level layers like rooms and buildings using the methods proposed in~\citep{hughes2022hydra}.

\paragraph{Scene Dynamics} To represent the changing state of the scene, we define the \textit{event mapping} $\mathcal{E}$ as the function $(\mathbf{A}_i, t) \mapsto \Delta\theta,$ which maps an articulate object and timestamp to an \textit{articulation state}, which represents how much the articulated part has been articulated relative to its initial position (meters for prismatic, radians for revolute). Note that this formulation allows us to model the dynamics of constrained objects by propagating the state change along the edge from its parent object. For example, a drawer that is translated an amount $\Delta\theta$ along its axis $\mathbf{u}$ also causes all of its \textit{constrained} instances to translate the same amount.

\subsection{Estimating Articulation Models from Egocentric Data}
\label{sec:articulation-est}

We approach the problem of articulation model tracking and estimation as three subproblems: (1) identifying relevant interaction sequences; (2) estimating joint parameters for newly discovered objects; and (3) tracking articulation states of these objects when they are revisited. This allows us to estimate $\mathcal{A}, \mathcal{P}, \textrm{and } \mathcal{E}.$

\paragraph{Keyframe Detection}
\label{sec:keyframe-detection}

\begin{figure}[t]
  \centering
  \begin{minipage}[t]{.3\linewidth}
    \centering
    \includegraphics[width=\linewidth]{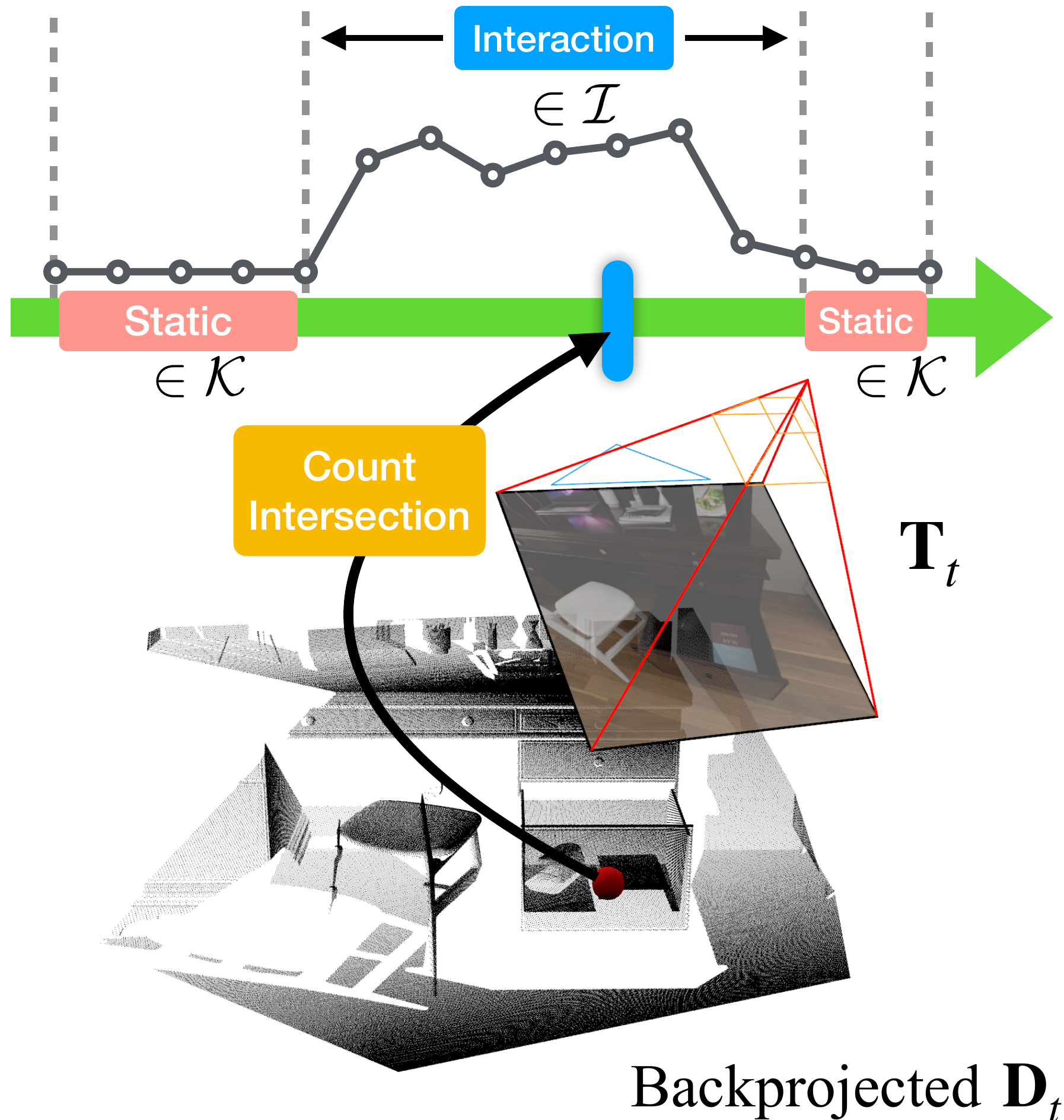}
  \end{minipage}
  \hfill
  \begin{minipage}[t]{.68\linewidth}
    \centering
    \includegraphics[width=\linewidth]{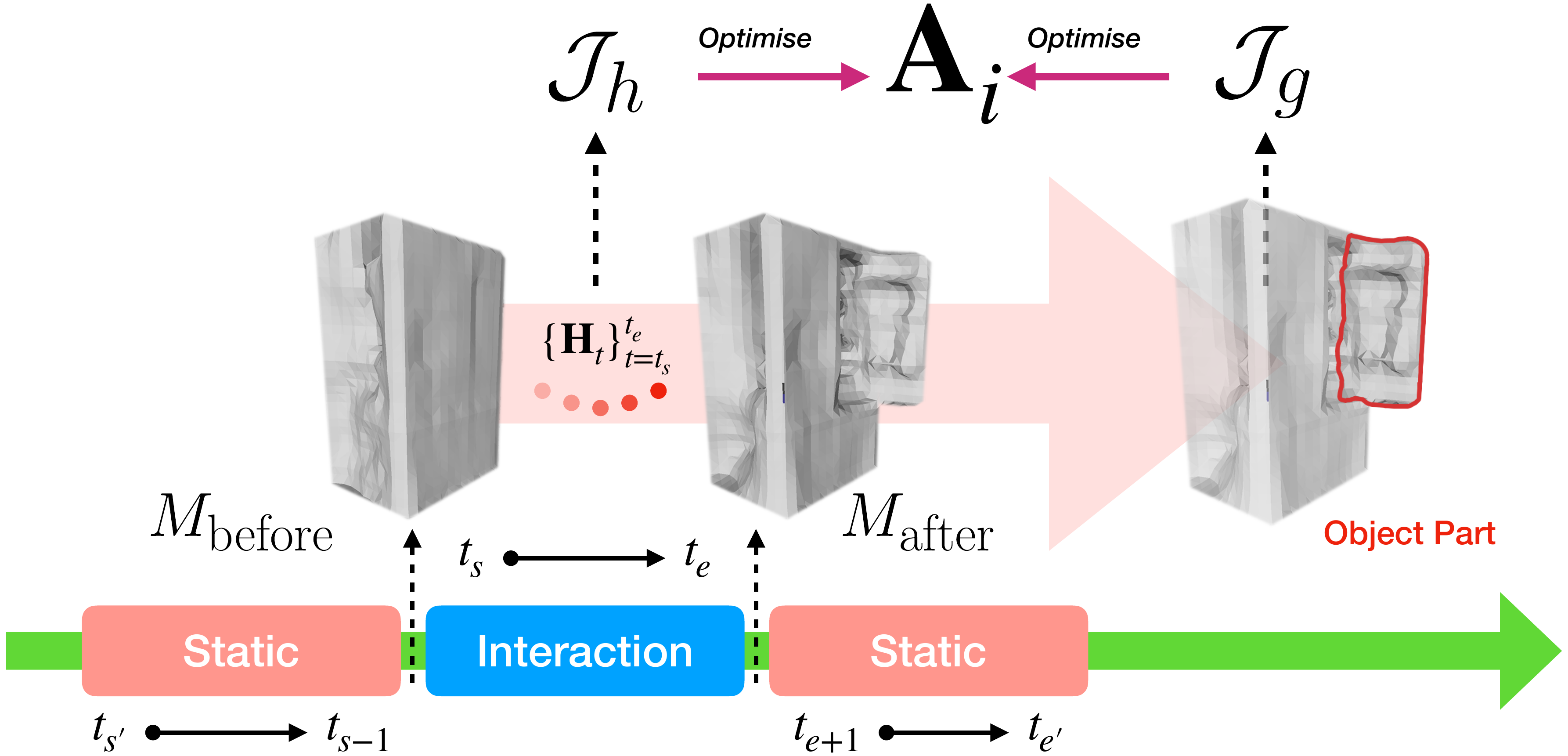}
  \end{minipage}
  \caption{\textbf{(left)} For each timestep $t,$ we intersect the hand-sphere with the scene pointcloud (backprojected from the depth map $\mathbf{D}_t$). The counts form a time series, which we post-process to obtain the interaction interval, with endpoints that define the \textit{interaction keyframes} $\in \mathcal{I}.$ The \textit{static keyframes} $\in \mathcal{K},$ where there is no interaction,  span the intervals in red.
           \textbf{(right)} From fusing the scene mesh during the \textit{static} keyframe intervals, we obtain the mesh before and after interaction. The object part is then extracted and used in combination with the hand poses to estimate the articulation model.}
  \label{fig:keyframe-articulation}
\end{figure}
As shown in Figure~\ref{fig:keyframe-articulation}, we begin by computing a set of \textit{interaction keyframes} $\mathcal{I},$ which consists of non-overlapping tuples representing the start and end timestamps of an interaction interval. We define the complement of $\mathcal{I}$ as the set $\mathcal{K}$ of \textit{static keyframes}, where each element contains the start and end timestamps of a period without interaction. At each timestamp $t,$ we represent the hand as a ball $\mathcal{B}_{R_\textrm{hand}}(\mathbf{H}_t)$ of radius $R_{\textrm{hand}},$ and backproject the depth $\mathbf{D}_t$ into a pointcloud.
We then compute a binary indicator
$
b_t =
\mathbbm{1}\!\bigl\{
\bigl|\,\mathcal{B}_{R_{\text{hand}}}(\mathbf{H}_t)\,\cap\,\mathbf{D}_t\,\bigr| > \delta \bigr\},
$
which indicates whether at least $\delta$ points of the (back-projected) point cloud
$\mathbf{D}_t$ fall inside
\(\mathcal{B}_{R_{\text{hand}}}(\mathbf{H}_t)\) (intuitively, this is the case where the user's hand is touching some object in the scene).
The resulting binary time-series $\{b_t\}$ is temporally smoothed by opening and closing morphological filters, with a kernel size corresponding to one second. $\mathcal{I}$ and $\mathcal{K}$ are then computed by partitioning $\{b_t\}$ into blocks of ones and zeros, respectively. Knowing $\mathcal{I}$ allows us to focus on the hand positions $\mathbf{H}_t$ during interactions, which will inform articulation model estimation. At each interaction interval $(t_s, t_e) \in \mathcal{I}$, we either \textit{discover} a new object, in which case we estimate its joint parameters and state, or \textit{update} an existing object, where we estimate how much the object was articulated from its previous state, as described below.

\paragraph{Discovering New Objects}
\label{sec:articulation-estimation}
We start by finding an estimate of $\mathbf{P}_i$ for the geometry of the object part, as illustrated in Figure~\ref{fig:keyframe-articulation}.
 Let $(t_{s'}, t_{s-1}), (t_{e+1}, t_{e'}) \in \mathcal{K}$ be the static keyframe pairs that surround the interaction. We fuse a mesh for both intervals using a pre-trained Egocentric Voxel Lifting (EVL) model~\citep{straub2024efm3d} to give $M_{\textrm{before}}$ and $M_{\textrm{after}},$ respectively. To extract $\mathbf{P}_i,$ we initialize a flood-fill at $\mathbf{H}_{t_e}$ on the vertices of $M_{\textrm{after}},$ and terminate when a vertex of $M_\textrm{before}$ is reached. Next, we estimate the part's articulation type $\kappa_i$, joint parameters $\mathbf{A}_i,$ and state $\Delta\theta.$ Let $\mathcal{H} = \{\mathbf{H}_t\}_{t=t_s}^{t_e}$ be the hand poses during the interaction. We fit a circular arc and a line to $\mathcal{H},$ and determine $\hat\kappa$ based on the minimum fit residual. We use the fit parameters as initial guesses $\mathbf{A}_{i, 0}, \Delta\theta_0.$ To determine $\mathbf{A}_i$ and $\Delta\theta,$ we optimize a cost function with two terms (and a regularizer). The first term $\mathcal{J}_g$ penalizes how far the vertices of the object, if articulated back to their original state, deviate from the scene mesh before interaction. More concretely, let $\mathbf{T}^{(\kappa_i)}(\cdot, \beta; \mathbf{A}_i, \Delta\theta )$ denote the 1D rigid motion operator by an amount $-\beta\Delta\theta,$ for $\beta \in [0, 1].$ Then
\begin{equation}
    \mathcal{J}_g (\mathbf{A}_i, \Delta\theta) = \frac{1}{|\mathbf{P}|} \sum_{\mathbf{p} \in \mathbf{P}}\left|\left|\mathbf{T}^{(\kappa_i)}(\mathbf{p}, 1; \mathbf{A}_i, \Delta\theta) - \mathbf I(\mathbf{p}, M_\textrm{before})\right|\right|_2^2,
\end{equation}
where $\mathbf{I}(\mathbf{p}, M_\textrm{before}) = \text{argmin}_{\mathbf{q} \in M_\textrm{before}} ||\mathbf{p} - \mathbf{q}||_2.$ In other words, $\mathcal{J}_g$ is the one-sided Chamfer distance between the estimated object points before interaction and the scene mesh. The second term $\mathcal{J}_h$ considers consistency with the hand trajectory. Let $\mathbf{h}_j \in \mathcal{H}, 1 \leq j \leq H$ be the chronologically ordered hand positions, and let $\alpha_j \in [0, 1]$ satisfy $\alpha_1 = 0, \alpha_H = 1.$ Then, $\mathcal{J}_h  $ encourages all hand points to align with the initial hand pose $\mathbf{h}_1$ when transformed by $\mathbf{T}^{(\kappa_i)}(\cdot, \alpha_j; \mathbf{A}_i, \Delta\theta).$ Concretely,
\begin{equation}
    \mathcal{J}_h(\mathbf{A}_i, \Delta \theta, \alpha) = \frac{1}{H}\sum_{j=1}^H\left|\left|\mathbf{T}^{(\kappa_i)}(\mathbf{h}_i, \alpha_j;\mathbf{A}_i, \Delta\theta) - \mathbf{h}_1\right|\right|_2^2.
\end{equation}
Overall, we solve
\begin{equation}
    \underset{\mathbf{A}_i, \Delta \theta, \{\alpha_j\}_{j=1}^H}{\min} \lambda_g\mathcal{J}_g(\mathbf{A}_i, \Delta\theta) + \lambda_h \mathcal{J}_h(\mathbf{A}_i, \Delta\theta, \alpha) + \lambda_p(\left|\left|\mathbf{A}_i-\mathbf{A}_{i,0}\right|\right| + |\Delta\theta-\Delta\theta_0|),
\end{equation}
where $\lambda_g, \lambda_h$ are weights controlling the costs, and $\lambda_p$ is the regularization weight. We then populate $\mathcal{A}$ and $\mathcal{E}$ with the solutions $\mathbf{A}_i$ and $\Delta\theta,$ respectively. Moreover, we update the set $\mathcal{P}$ with $\mathbf{P}_i.$ Note that we do not model $\mathcal{E}$ within $(t_s, t_e).$ However, since we assume the scene is static outside of $\mathcal{I},$ estimating $\mathcal{E}$ this way addresses all times in $\mathcal{K}.$

\paragraph{Updating Existing Objects}
\label{sec:updating-objects}
To determine whether the user is interacting with an object that has already been discovered, we consider the hand-ball $\mathcal{B}_{R_\textrm{hand}}(\mathbf{H}_{t_s})$ at the start of the interaction. Then, we check if $\max_{\mathbf{P}_i \in \mathcal{P}} \frac{|\mathcal{B}_{R_\textrm{hand}}(\mathbf{H}_t) \cap \mathbf{P}_i|}{|\mathbf{P}_i|} > \sigma,$ and take the maximizer as the revisited object. If no such object is found, this is considered a new object, and we estimate its parameters as previously described. Otherwise, we let $\mathbf{P}_i, \mathbf{A}_i, \Delta\theta$ be the geometry, articulation parameters, and articulation state of the identified object. To update the revisited object's articulation state from hand poses $\mathcal{H},$ we iteratively transform $\mathbf{P}_i$ to get the updated pointcloud $\mathbf{P}_i'$ and articulation state $\Delta\theta',$ as outlined in Algorithm~\ref{alg:articulation-updating} in Appendix~\ref{appendix:updating-existing-objects}. Specifically, we treat the first hand pose $\mathbf{h}_1 \in \mathcal{H}$ as an anchor point and sequentially update $\mathbf{P}_i'$ with each $\mathbf{h}_j \in \mathcal{H},$ if the hand-intersection check $\frac{|\mathcal{B}_{R_\textrm{hand}}(\mathbf{h}_i) \cap \mathbf{P}_i'|}{|\mathbf{P}_i'|} > \sigma$ is satisfied. To get the articulation amount between $\mathbf{h}_i$ and $\mathbf{h}_1,$ we assume the entire object transforms with the anchor point following $\mathbf{A}_i$. For example, a prismatic object translates by an articulation amount of $\varphi = \langle \mathbf{h}_i - \mathbf{h_1}, \mathbf{u}\rangle,$ where $\mathbf{u}$ is the prismatic axis. We then update $\mathcal{P}$ and $\mathcal{E}$ with $\mathbf{P}_i'$ and $\Delta\theta + \varphi$ respectively.

\subsection{Articulated 3D Scene Graph Construction}
\label{sec:method-sg}
We use the estimates of $\mathcal{A}, \mathcal{P}, \mathcal{E}$ to construct $\mathcal{G}$ from $\{\mathbf{S_t}\}.$ For each $(t_s, t_e) \in \mathcal{K},$ we use the sequence $\{\mathbf{S}_t\}_{t=t_s}^{t_e}$ of observations to update $\mathcal{G}.$ Let $(t_s', t_e')\in \mathcal{I}$ be the interaction interval that precedes $(t_s, t_e),$ and let $\mathbf{P}_i$ and $\mathbf{A}_i$ be the pointcloud and articulation parameters of the articulate object. Each sequence of observations can follow (1) a \textit{static} interval, where there was no interaction; (2) a \textit{discovery} interval, where a new articulate object was discovered; or (3) an \textit{update} interval, where there was interaction with a known object. To integrate $\{\mathbf{S}_t\}_{t=t_s}^{t_e}$ into $\mathcal{G},$ we build on the memory and perception modules of RoboEXP~\citep{jiang2024roboexp}, where each $\mathbf{I}_t$ is processed by GroundingDINO~\citep{liu2023grounding}, SAM-HQ~\citep{sam_hq}, and per-instance CLIP~\citep{radford2021learningtransferablevisualmodels}. Together, these convert the observation into a set of segmented pointclouds annotated with a CLIP feature, which we refer to as $\mathbf{O}.$ In all cases, we update $\mathcal{G}$ by backprojecting $\mathbf{O}$ into the \textit{Geometric Layer}'s voxel grid. All object instances are then registered in the \textit{Object Layer} and connected to the voxels they consist of with an interlayer edge. Lastly, overlapping instances are merged by CLIP similarity.

The three cases differ by how we filter $\mathbf{O}$ before backprojecting into the voxel grid. The key idea is that \textit{contained} objects are merely occluded by the articulate object before interaction, and that there should be geometric overlap between \textit{constrained} objects and the articulate object after interaction. In \textit{static} intervals, no filtering is involved. For \textit{discovery} intervals, run the process twice to check for \textit{contains} and \textit{constrains} separately. We compute $\mathbf{P}_{i,\textrm{before}}$ by articulating $\mathbf{P}_i$ the amount $\mathcal{E}(\mathbf{A}_i, t_s')$ to get the points before interaction. We then only consider the subset of $\mathbf{O}$ that lie along rays occluded by $\textrm{conv}(\mathbf{P}_{i,\textrm{before}}),$ where $\textrm{conv}(\cdot)$ denotes the convex hull. After merging overlapping objects, we connect any newly discovered objects to the articulate part via a \textit{contains} edge. We then check for \textit{constrains} relationships by articulating $\mathbf{P}_i$ the amount $\mathcal{E}(\mathbf{A}_i, t_s')$ to obtain $\mathbf{P}_{i,\textrm{after}},$ but only consider the subset of $\mathbf{O}$ that overlaps with $\textrm{conv}(\mathbf{P}_{i,\textrm{after}}).$ By our assumptions, these discovered edges are static, so we lock them into the \textit{Objects Layer} so that they are not affected by future updates. We finish the \textit{discovery} case by pruning all voxels and objects that overlap with $\textrm{conv}(\mathbf{P}_{i,\textrm{before}}).$

Lastly, the \textit{update} case is treated the same as the \textit{static} case, except we also prune all voxels and objects that overlap with $\mathbf{P}_{i,\textrm{before}}.$ We combine the voxel grid with all pointclouds corresponding to dynamic objects (i.e., articulate parts and their constrained objects) to form the final \textit{Geometric Layer}.

\section{Experiments}

In Section~\ref{subsec:articulation-estimation} and \ref{subsec:scene-graph-reconstruction}, we evaluate Pandora's articulated 3D scene graph quantitatively from simulated data against baseline methods. We create environments in Blender to emulate the data collection process in the real world, as discussed in Appendix~\ref{appendix:articulation-baselines}. In Section~\ref{subsec:real-world-results} and \ref{subsec:robotics-application}, we evaluate Pandora in real-world scenes and demonstrate its downstream capabilities for object-retrieval.

\subsection{Articulation Model Estimation}
\label{subsec:articulation-estimation}
\paragraph{Metrics} For a given object, let $\mathbf{u}^*$ be the (normalized) ground truth axis direction and $\mathbf{\hat u}$ be the (normalized) estimate. Similarly, let $\kappa^*, \hat{\kappa} \in \{\mathrm{Prismatic, Revolute}\}.$ We measure the axis error in terms of the cosine distance as
\begin{equation}
    \mathcal{L}_{\textrm{axis}}(\mathbf{u}^{*},\hat{\mathbf{u}}, \kappa^*, \hat\kappa) = W_{\pi}\!\bigl(\cos^{-1}\langle\mathbf{u}^{*},\mathbf{\hat{u}}\rangle\bigr) \cdot \mathbbm{1}(\hat\kappa = \kappa^*) + \frac{\pi}{2} \cdot \mathbbm{1}(\hat\kappa \neq \kappa^*),
\end{equation}
where $W_\pi(\theta) = \min(\theta, \pi -\theta)$ is a wrapping function. For the revolute pivot point, we consider a \textit{normalized} distance with respect to a canonical length of the link. Concretely, let $d$ be the length of the bounding box diagonal of the object's articulate part, and let $(\mathbf{x}^*$, $\mathbf{\hat x})$ be the pivot ground truth and estimate, respectively. We then compute
\begin{equation}
  \mathcal{L}_{\textrm{pivot}}
  (\mathbf{u}^*,\mathbf{\hat u},\mathbf{x}^*,\mathbf{\hat x}, \kappa^*, \hat\kappa)=
    \frac{\|(\mathbf{x}^*-\mathbf{\hat x})\times\mathbf{u}^*\|_2}{d} \cdot \mathbbm{1}(\hat\kappa = \kappa^*)
        + 1 \cdot \mathbbm{1}(\hat\kappa \neq \kappa^*).
\end{equation}

\paragraph{Baselines}
\label{sec:experiments-articulation-estimation}
We compare our method with state-of-the-art articulation model estimators from different input modalities. Articulate Anything~\citep{le2024articulate} takes a short RGB-only video of object interaction and aims to reconstruct the target articulation model by using a VLM to propose and critique link and joint placements. Ditto~\citep{jiang2022dittobuildingdigitaltwins} takes in segmented pointclouds before and after interaction with the object and infers the joint parameters decomposes it into a static and mobile part. Our Hands-Only baseline runs the same optimization described in Sec.~\ref{sec:articulation-est} but only using hand pose information (e.g., random joint initialization, with $\lambda_g = \lambda_p = 0$).

\paragraph{Results} We present results in Table~\ref{tab:articulation-estimation} and show that egocentric data allows us to achieve competitive performance with our model-based approach against state-of-the-art alternatives. Data from human-object interactions naturally provides more information about articulation and is straightforward to collect with the Aria device~\citep{engel2023projectarianewtool}.
We provide an analysis of these baselines and an explanation for the apparent zero prismatic axis error observed for Articulate Anything in Appendix~\ref{appendix:articulation-baselines}. Our full method performs well overall, but relies on accurate tracking of the hand poses; in real-world scenes, occasional inaccuracies can affect the joint estimation. We expect this to improve as hand pose estimation becomes more precise.

\subsection{Articulated Scene Graph Reconstruction}
\label{subsec:scene-graph-reconstruction}

\begin{wraptable}[11]{R}{.48\linewidth}
  \vspace{-1.5\baselineskip}
  \centering
  \footnotesize
  \caption{Comparison of methods on prismatic and revolute joints.  Axis errors presented in degrees; pivot error normalized by each
  object's bounding-box diagonal. }
  \label{tab:articulation-estimation}
  \setlength{\tabcolsep}{3pt}
  \renewcommand{\arraystretch}{1.05}
  \begin{tabular}{l l c c c}
    \toprule
    \multicolumn{2}{c}{\textbf{Method}} &
      \textbf{Prism.} & \multicolumn{2}{c}{\textbf{Revol.}} \\
    \cmidrule(lr){1-2}\cmidrule(lr){3-3}\cmidrule(lr){4-5}
    \textbf{Name} & \textbf{Input} &
      \textbf{Axis$\downarrow$} & \textbf{Axis$\downarrow$} & \textbf{Pivot$\downarrow$} \\
    \midrule
    Art.\ Anything & RGB Seq.   & 0.00 & 0.00 & 0.22 \\
    Ditto               & 2 PCs      & 10.82 & 1.26 & 0.26 \\
    Hands-Only          & Hand Seq.  & 0.02 & 15.14 & 0.63 \\
    Ours                & Ego Seq.   & 0.19 & 0.62 & 0.004 \\
    \bottomrule
  \end{tabular}
\end{wraptable}
\paragraph{Metrics} We evaluate Pandora's ability to model the \textit{Geometric Layer} and \textit{Object Layer} of the articulated 3D scene graph. To better compare across methods, we reduce ambiguity in the detected objects by replacing the Grounding DINO + SAM-HQ + CLIP processing pipeline from Section~\ref{sec:method-sg} with ground-truth instance maps. Similar to~\citep{Schmid-RSS24-Khronos}, we focus on the following metrics, and report the precision, recall, and F1 score of each: (1) \textbf{background mesh recovery}, where a vertex is considered positive if there is a ground truth vertex within $\delta=20$cm of it; (2) \textbf{static objects}, where an object is a positive if its bounding box overlaps with a corresponding ground truth static object with the same instance label; (3) \textbf{dynamic objects}, where an object is positive at a \textit{keyframe} $\in \mathcal{K}$ if its centroid is within $\delta=10$cm of a corresponding ground truth dynamic object with the same instance label; and (4) \textbf{object relationships}, where an object is positive if it is linked to the correct ground truth container object with the appropriate relationship (\textit{contains} or \textit{constrains}). Note that for dynamic objects, we average each metric over the set of keyframes $\mathcal{K}$. For simplicity, we only model one level of the ``contains'' relationship (i.e., no nested containers).

\paragraph{Baselines} We evaluate against (1) Hydra~\citep{hughes2022hydra}, a real time, globally consistent spatial perception system that reconstructs the environment as a hierarchial 3D scene graph; and (2) Khronos~\citep{Schmid-RSS24-Khronos}, a system for spatio-temoral metric-semantic SLAM that represents the environment as a dense background mesh and object fragments. Note that neither method models object relationships, so we additionally compare against our own method without using the estimated articulate part geometry (NoArt) to inform which objects are related to it.

\paragraph{Results} We present our results in Table~\ref{tab:experiment-sg}. Because Pandora explicitly detects articulate object parts, we are able to on average isolate the background mesh more precisely than the baseline methods. Hydra~\citep{hughes2022hydra} is intended for static scenes and does not account for any dynamics, but picks up on some static background objects. Khronos~\citep{Schmid-RSS24-Khronos} is able to detect some dynamic objects correctly but struggles because it only models object centroid translation, and is not aware that parts of an object can move independently. In contrast, since we model the \textit{constrains} relationship, our method can relate a dynamic object's movement to its parent object kinematics and improve its estimated tracks. Furthermore, from NoArt, we see that knowledge of articulation models is important for understanding relationships.

\begin{table}[t]
  \centering
  \footnotesize
  \setlength{\tabcolsep}{3pt}
  \renewcommand{\arraystretch}{1.0}

  \caption{Articulated 3D scene graph reconstruction metrics in simulated environments. Unlike~\citep{Schmid-RSS24-Khronos}, we only focus on dynamic objects which are constrained to move with an articulate part. NoArt baseline is ours without using the estimated articulate part geometry to determine object relationships.}
  \label{tab:experiment-sg}

  \begin{tabular}{@{}L{1.6cm}L{1.9cm}cccc@{}}
    \toprule
    \multirow{2}{*}{Dataset} & \multirow{2}{*}{Method} &
      \multicolumn{4}{c}{\textbf{Metric (Pre / Rec / F1) $\uparrow$}}\\
    \cmidrule(l){3-6}
      & & Mesh Recovery & Static Objects & Dynamic Objects & Object Relationships\\
    \midrule
    \multirow{4}{*}{Kitchen}
      & Hydra~\citep{hughes2022hydra}              & 88.3 / \textbf{50.6} / 64.3 & \textbf{75.0} / 62.1 / 68.0 & -- & -- \\
      & Khronos~\citep{Schmid-RSS24-Khronos}       & 92.7 / \textbf{50.6} / \textbf{65.4} & 66.7 / 46.7 / 54.9 & 16.8 / 6.9 / 9.8 & -- \\
      & NoArt                                     & 98.5 / 46.0 / 63.0 & 57.1 / \textbf{93.3} / 70.9 & -- & 30.7 / 50.0 / 41.2 \\
      & \textbf{Ours}                             & \textbf{98.6} / 46.0 / 62.7 & \textbf{75.0} / 90.0 / \textbf{81.8} & \textbf{100.0} / \textbf{91.7} / \textbf{95.7} & \textbf{100.0} / \textbf{100.0} / \textbf{100.0} \\
    \midrule
    \multirow{4}{*}{Bedroom}
      & Hydra~\citep{hughes2022hydra}              & 90.1 / \textbf{68.8} / 78.0 & \textbf{86.7} / 60.5 / 71.2 & -- & -- \\
      & Khronos~\citep{Schmid-RSS24-Khronos}       & \textbf{98.2} / 68.3 / \textbf{80.6} & 42.6 / 60.5 / 50.0 & 1.1 / 0.7 / 0.8 & -- \\
      & NoArt                                     & 98.1 / 66.2 / 79.0 & 61.5 / \textbf{93.0} / 74.1 & -- & 41.0 / 60.0 / 48.7 \\
      & \textbf{Ours}                             & 98.1 / 66.2 / 79.0 & 72.7 / \textbf{93.0} / \textbf{81.6} & \textbf{93.5} / \textbf{84.1} / \textbf{88.5} & \textbf{87.5} / \textbf{93.3} / \textbf{90.3} \\
    \bottomrule
  \end{tabular}
\end{table}

\subsection{Real World Results}
\label{subsec:real-world-results}

We collect two real-world kitchen datasets with the Aria device to evaluate Pandora's articulation estimation and object retrieval accuracy, presented in Table~\ref{tab:real-results}. We manually annotate the articulation parameters of objects in the scene and compute the same metrics from Section~\ref{subsec:articulation-estimation}. Toward measuring object retrieval accuracy, we consider a fixed set of concealed objects and mark a detection successful if the object is given the correct semantic label and associated with the correct parent object. We observe that the most common failure case is when the geometry of the articulate part is inaccurately estimated, resulting in missing detections.

\subsection{Robotics Application: Object Retrieval}
\label{subsec:robotics-application}

\begin{figure}[t]
  \centering
  \includegraphics[width=0.7\linewidth]{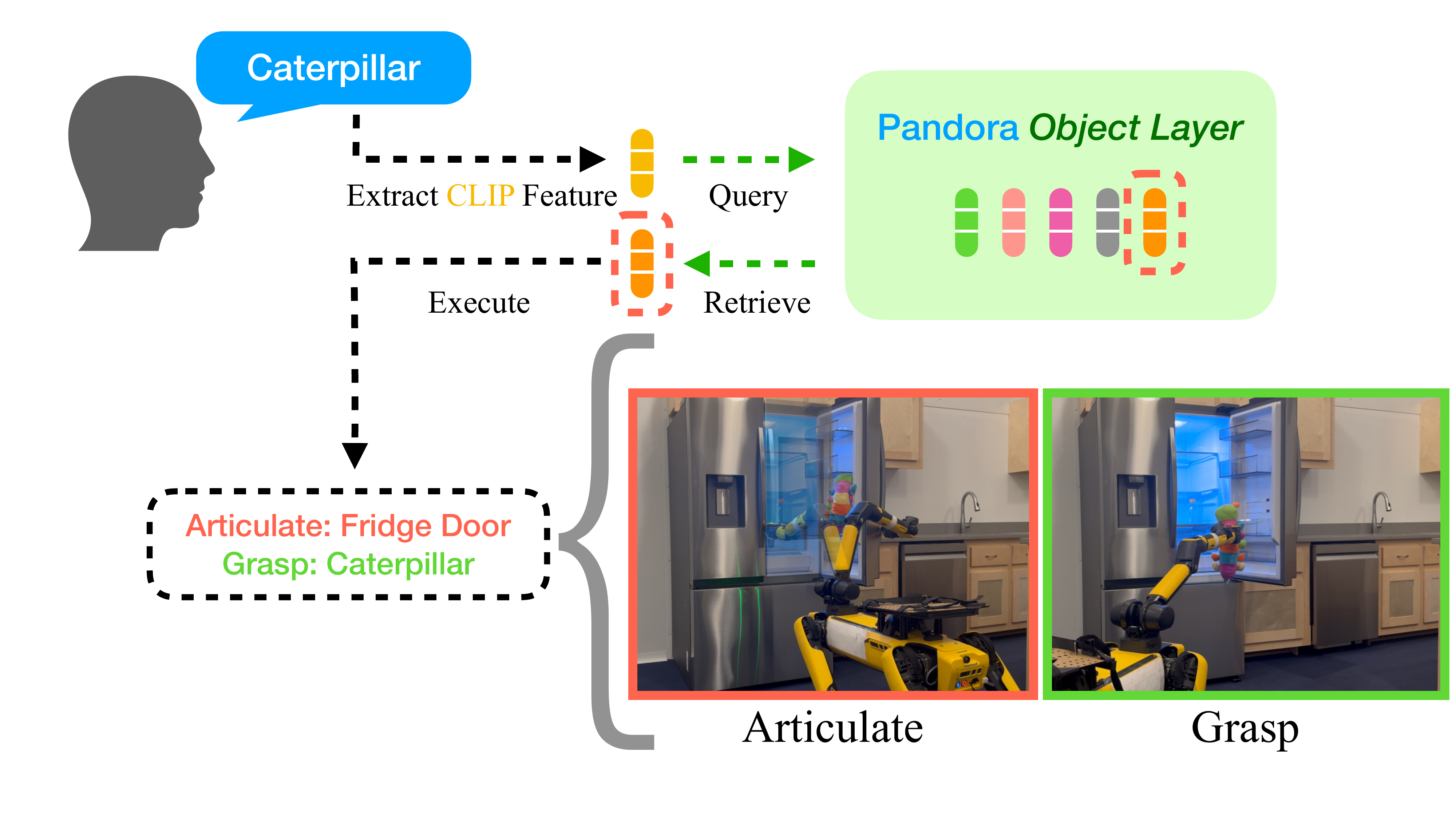}
  \caption{Pandora is queried to provide the handle grasp point and orientation, along with target end-effector positions. With knowledge of how the \textit{constrained} object has moved, a target grasp location is proposed and executed.}
  \label{fig:spot-object-retrieval}
\end{figure}
\label{sec:spot-object-retrieval}
In Figure~\ref{fig:spot-object-retrieval}, we present a demonstration of a Boston Dynamics Spot being tasked with retrieving a concealed object. A human scans the scene with Aria~\citep{engel2023projectarianewtool} glasses and reconstructs the articulated scene graph through Pandora. We then provide a language query $L$ as input and search the \textit{Object Layer} of Pandora for the object with the closest CLIP~\citep{radford2021learningtransferablevisualmodels} feature in cosine distance. We decompose the retrieval task into (1) navigating to the object's parent object node, (2) grasping the articulate part's handle, (3) articulating the part using its estimated joint parameters, and (4) grasping the target object centroid. All navigation and grasp targets are computed inside Pandora. We provide additional details about the robot execution setup in Appendix~\ref{appendix:spot-execution} and refer the reader to the supplemental video.

\section{Conclusion}

\begin{table}[t]
  \centering
  \footnotesize
  \caption{Results from Pandora on manually annotated real world scenes. The articulation metrics are computed as in Table~\ref{tab:articulation-estimation} and  Section~\ref{subsec:articulation-estimation}. Object retrieval is successful if the object is detected and associated with the correct parent object and semantic label. Axis errors are presented in degrees.}
  \label{tab:real-results}
  \setlength{\tabcolsep}{6pt}
  \renewcommand{\arraystretch}{1.05}
  \begin{tabular}{l c c c c}
    \toprule
    \textbf{Dataset} & \textbf{Prism.\ Axis$\downarrow$} & \textbf{Revol.\ Axis$\downarrow$} & \textbf{Revol.\ Pivot$\downarrow$} & \textbf{Retrieval Acc$\uparrow$} \\
    \midrule
    Kitchen A & 3.76 & 3.01 & 0.10 & 5/6 \\
    Kitchen B & 2.77 & 4.30 & 0.12 & 5/6 \\
    \bottomrule
  \end{tabular}
\end{table}

We present Pandora, a system for creating \textit{articulated} 3D scene graphs from egocentric data which can be used by a robot for object-retrieval tasks. Delegating the mapping step to a human leads to a more \textit{complete} scene graph, rather than being constrained by a robot mapper's skills or embodiment. We show that egocentric vision provides a rich yet convenient source of data, enabling articulation model estimation with simple heuristics. Pandora recovers scene geometry, static objects, dynamic object tracks, and object-container relationships.

\paragraph{Limitations} Pandora does not handle cases with partial observations very well. Specifically, if the entire object part is not observed, $\mathcal{J}_g$ may hurt the articulation model estimate. Leveraging semantic priors about how objects might be articulated (similar to~\citep{le2024articulate}), could be helpful for addressing this issue. To better focus on our contributions, we also restricted the class of dynamic objects to those that are constrained to articulate parts. It would be desirable in future work to expand on this and track arbitrary short and long-term changes (as studied in~\citep{Schmid-RSS24-Khronos}).

\acknowledgments{This work was partially supported by Meta Reality Labs through the Project Aria research program, and by the ARL DCIST CRA under grant W911NF-17-2-0181. We gratefully acknowledge the Project Aria team for providing hardware, tools, and support essential to this project.}

\bibliography{refs}

\clearpage

\appendix
\section{Inferring Depth from the Aria Device}
\label{appendix:inferring-depth}
Since the Aria device currently does not collect depth images, we estimate the depth maps $\mathbf{D}_t \in \mathbb{R}^{H\times W}$ for each timestamp. We do this by using a monocular depth estimator on the RGB image $\mathbf{I}_t,$ and then using an estimated metric surface to extract scale parameters.

The Aria device provides two mono-SLAM streams as $\mathbf{M}_t \in \mathbb{R}^{2 \times H\times W},$ which makes it possible to extract a semidense pointcloud $\mathbf{P}$ from the Project Aria Machine Perception Services (MPS)~\citep{engel2023projectarianewtool}. We then chunk the timestamps $\{1, \cdots, N\}$ into $K$ blocks of size $\Delta= \lceil \frac{N}{K}\rceil.$ For each $k \in [K],$ we provide the stream $\{\mathbf{M}_t, \mathbf{I}_t, \mathbf{T}_t\}_{t=k\Delta}^{(k+1)\Delta}$ along with $\mathbf{P}$ into a pretrained Egocentric Voxel Lifting (EVL) model~\citep{straub2024efm3d}, which estimates a surface mesh of the scene. Chunking is important because EVL only estimates static meshes, and we cannot assume apriori knowledge of when interactions occur. For each $t,$ we compute normalized disparity maps $\hat{\mathbf{D}}_t$ by passing in $\mathbf{I}_t$ to a monocular depth estimator (Depth Anything V2~\citep{depth_anything_v1, depth_anything_v2}). We also render the metric depth $\tilde{\mathbf{D}}_t$ from the corresponding camera pose. We then sample a subset of pixels $\mathcal{P}$ to regress the scale and shift parameters as
\begin{equation}
    s^*(t), s_0^*(t) =\arg\min_{s, s_0} \sum_{p \in \mathcal{P}}\left|\frac{s}{\hat{\mathbf{D}_t}(p) + s_0} -\tilde{\mathbf{D}}_t(p)\right|,
\end{equation}
and save the depth estimate as $$\mathbf{D}_t = \frac{s^*(t)}{\hat{\mathbf{D}_t}+s_0^*(t)}.$$

In practice, we use a refined version of $\mathcal{P}$ by also rendering a semantic mask from a segmentation network (OneFormer~\citep{jain2023oneformer}) and only considering \textit{background} classes from the ADE20k label set~\citep{zhou2017scene, zhou2019semantic} (e.g.\ wall, floor).

\section{Updating Articulation States}
\label{appendix:updating-existing-objects}
\begin{algorithm}[h]
  \caption{Updating Existing Objects}
  \label{alg:articulation-updating}
  \begin{algorithmic}[1]
    \Require Object Pointcloud $\mathbf{P}_i,$ Hand Poses $\{\mathbf{H}_t\}$, Articulation State $\Delta\theta$
    \State $\mathbf{h}_1 \gets \mathcal{H}[1]$
    \Comment{First hand pose as anchor}

    \State $\mathbf{P}_i' \gets \mathbf{P}_i$
    \State $\varphi \gets 0$
    \Comment{Relative to $\Delta\theta$}

    \For{$\mathbf{h}_j \in \mathcal{H}$}
        \If{$\frac{|\mathcal{B}_{R_\textrm{hand}}(\mathbf{h}_j) \cap \mathbf{P}_i'|}{|\mathbf{P}_i'|} > \sigma$}
            \State $\varphi \gets $ max(\Call{GetArticulationAmount}{$\mathbf{A}, \mathbf{h}_1, \mathbf{h}_j$}, $\varphi$)
            \State $\mathbf{P}_i' \gets \mathbf{T}^{(\kappa_i)}(\mathbf{P}_i, 1; \mathbf{A}_i, \varphi)$

        \EndIf
    \EndFor
    \State $\Delta\theta' \gets \Delta\theta + \varphi$
    \Ensure  Updated Pointcloud $\mathbf{P}_i',$ Articulation State $\Delta\theta'$
     \end{algorithmic}
\end{algorithm}

\section{Robot Execution Implementation Details}
\label{appendix:spot-execution}
We locate handles by passing ``handle'' as an additional tag to prompt Grounding DINO. Each handle is matched as a \textit{constrained} object by Pandora to its corresponding articulate object. We then use the centroid of the handle pointcloud as a grasp target, and the pointcloud orientation to determine the gripper rotation. For grasping the target object, we use Pandora's \textit{Geometric Layer} to compute a collision-free trajectory for the gripper. Specifically, we compute a standoff point to position the robot before grasping by considering ray bundles originating at the target object and finding the direction with maximum free-space. Since all planning occurs inside Pandora, the grasping precision is limited and suffers from odometric drift, but retrieval is typically successful when the grasp lands.

\section{Simulated Evaluation Details}
\label{appendix:articulation-baselines}

\begin{figure}[t]
  \centering
  \begin{minipage}[t]{.49\linewidth}
    \centering
    \includegraphics[width=\linewidth]{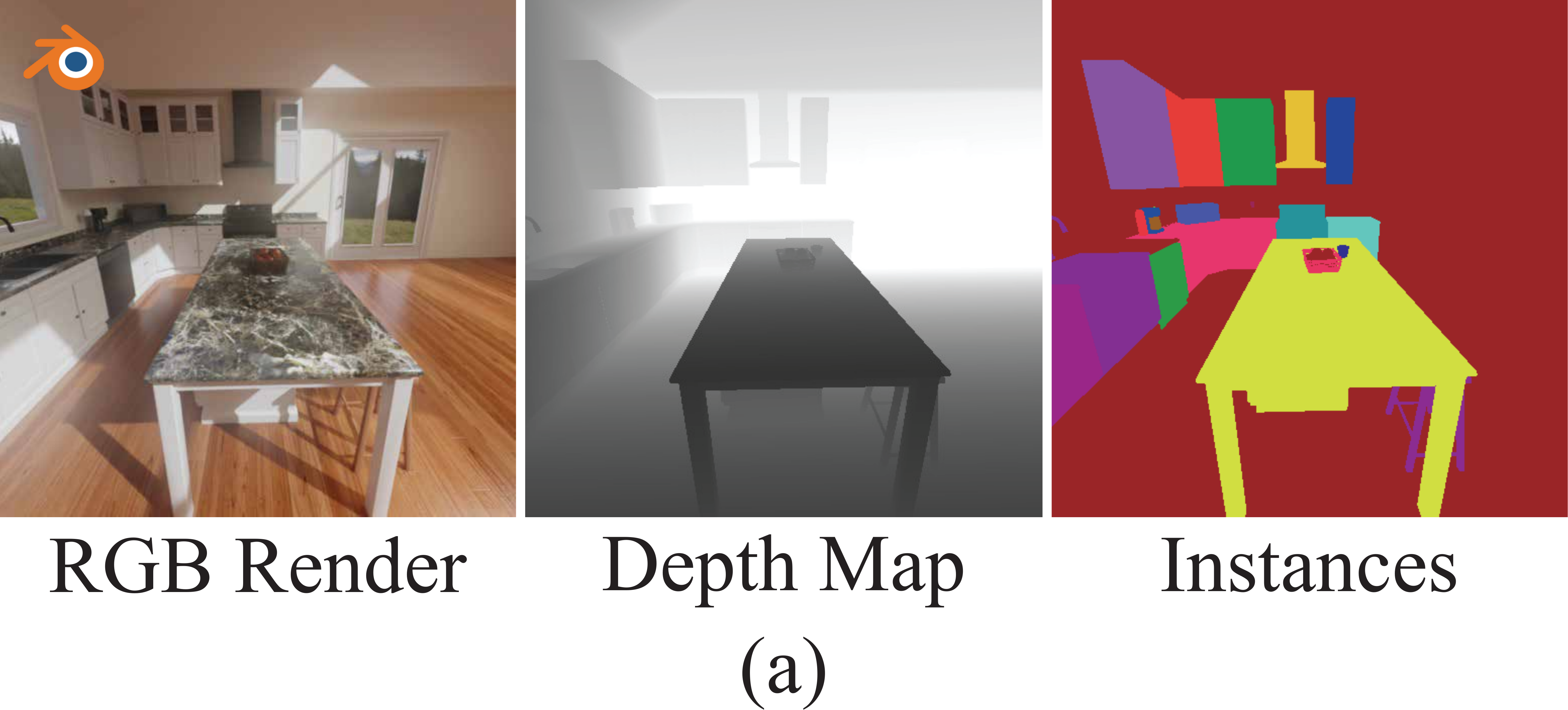}
  \end{minipage}
  \hfill
  \begin{minipage}[t]{.49\linewidth}
    \centering
    \includegraphics[width=\linewidth]{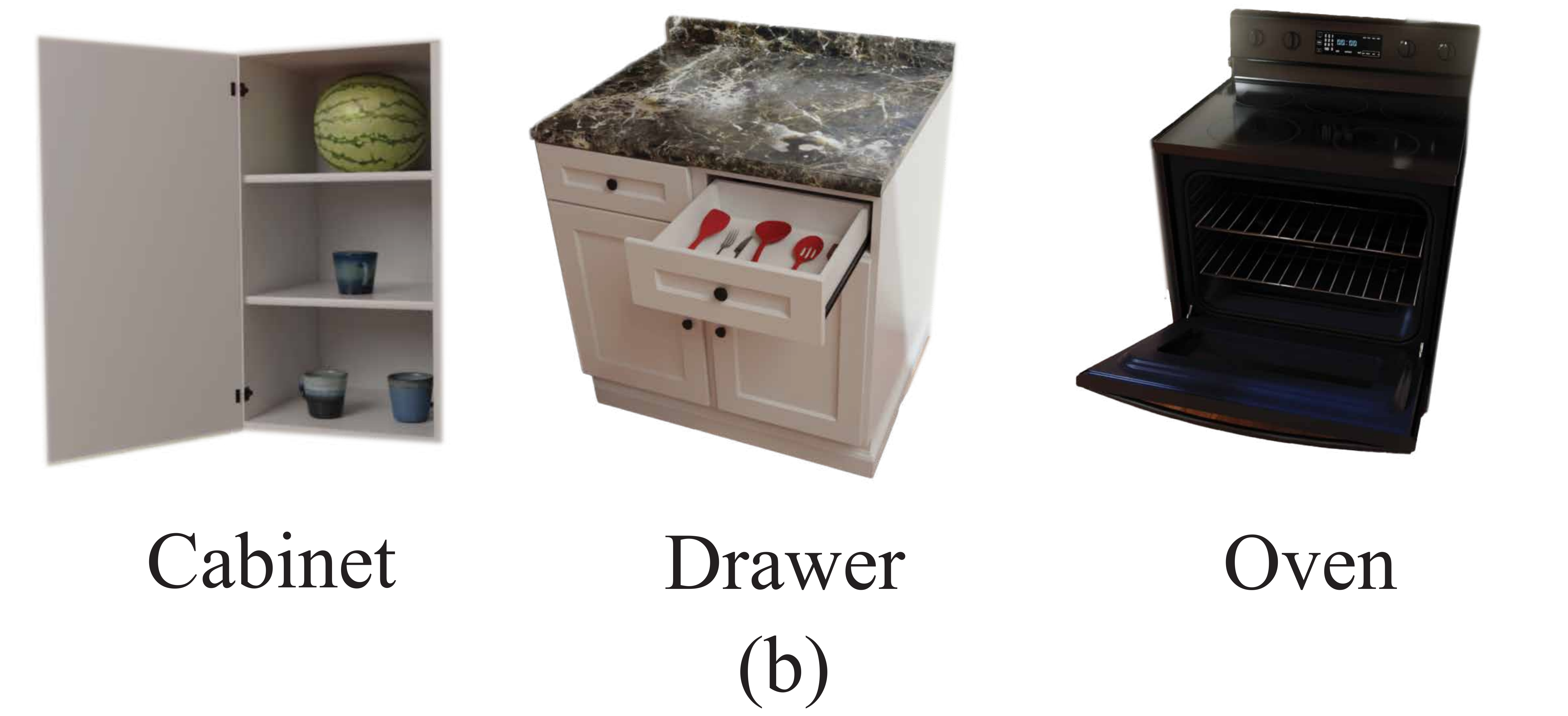}
  \end{minipage}
  \caption{(a) Sample renders from a simulated evaluation scene. The ground truth instance map is used for evaluations for consistency with baselines. (b) Examples of articulated objects in the evaluation scene. }
  \label{fig:blender-eval}
\end{figure}
\paragraph{Simulated Scenes} For each of our evaluation scenes, we set a camera trajectory and animate interactions with articulated objects in the environment. We then render RGB, depth, and instance maps (see Figure~\ref{fig:blender-eval}a in Appendix~\ref{appendix:articulation-baselines}). We label each articulate object with its joint parameters and provide an anchor point for the hand poses. We acknowledge existing realistic simulated environments as in~\citep{ai2thor, robocasa2024, adt} and human-object interaction datasets~\citep{kim2024parahome, grauman2022ego4dworld3000hours, Damen2018EPICKITCHENS}, but instead set up our own evaluation suite to better control the content (e.g., ``stocking'' refrigerators and drawers with test objects) and interactions (e.g., synthetic hand anchor positions, removing unwanted hand-object interactions) of each scene. We present sample renders from the Blender evaluation environments in Figure~\ref{fig:blender-eval}, along with examples of articulate objects in the scene.

\paragraph{Implementation Details for Articulation Model Baselines} Toward evaluating articulation estimation accuracy (Table~\ref{tab:articulation-estimation}), for each ground-truth articulate object, we label the keyframe pair $(s, t)$ of the start and end of interaction. We run Articulate Anything with Claude 3.5 Sonnet as the VLM and feed in the RGB render of each frame in $(s, t)$ as a video. We register the estimated URDF to the ground truth object mesh by aligning the estimated articulated link with that of the ground truth before computing the joint error. Ditto receives segmented object pointclouds at $s$ and $t.$ We observed that the released Ditto model often infers the incorrect articulation type, but sometimes still produces a reasonable result for the correct type. Hence, in our evaluation of Ditto, we only consider its joint prediction of the true articulation type, resulting in an upper bound on performance. For our methods, the simulated hand tracks are the trajectory of the labeled anchor point perturbed with random noise. As a proxy for the volumetric fusion step, we also take in the scene mesh at $s$ and $t.$ All assets are purchased from CGTrader or BlenderKit.

\paragraph{Articulation Model Baseline Results} Articulate Anything faces the same difficulties originally described by the authors in~\citep{le2024articulate}. In particular, the accuracy of the joint prediction is highly dependent on selecting the best object match from the PartNet-Mobility dataset~\citep{Mo_2019_CVPR, Xiang_2020_SAPIEN, chang2015shapenet} by appearance. It tends to successfully identify the joint type, resulting in very low axis error, but the proposed URDF often does not align with the ground truth object (e.g.\ doors swinging from the wrong side of the frame), resulting in high pivot errors. For the objects we tested on, Ditto tends to misclassify the joint type (we describe how we account for this in Appendix~\ref{appendix:articulation-baselines}), but mostly produces realistic results. As our optimization procedure is sensitive to the initial state, the Hands-Only baseline suffers as it lacks this information.

\end{document}